\pdfoutput=1

\documentclass[10pt,twocolumn]{article}

\usepackage[utf8]{inputenc}
\usepackage[T1]{fontenc}
\usepackage{amsmath,amssymb,amsfonts}
\usepackage{graphicx}
\usepackage{booktabs}
\usepackage{hyperref}
\usepackage{cite}
\usepackage{float}
\usepackage{tikz}
\usetikzlibrary{shapes.geometric, arrows.meta, positioning, fit, backgrounds, calc, decorations.pathreplacing}
\usepackage[margin=0.75in]{geometry}
\usepackage{abstract}
\usepackage{enumitem}
\usepackage{xcolor}
\usepackage{caption}
\usepackage{microtype}
\usepackage{lmodern}
\usepackage{authblk}

\definecolor{inputblue}{RGB}{200,220,255}
\definecolor{intentorange}{RGB}{255,235,210}
\definecolor{knowledgepurple}{RGB}{235,210,255}
\definecolor{responsegreen}{RGB}{210,245,225}
\definecolor{outputblue}{RGB}{210,230,250}
\definecolor{decisionyellow}{RGB}{255,245,180}
\definecolor{ibmblue}{RGB}{0,98,155}
\definecolor{linuxonegray}{RGB}{240,242,245}
\definecolor{spyreaccent}{RGB}{30,130,76}
\definecolor{watsonxpurple}{RGB}{105,50,160}

\hypersetup{
    colorlinks=true,
    linkcolor=ibmblue,
    citecolor=ibmblue,
    urlcolor=ibmblue
}

\title{\textbf{Spyre-Accelerated Retrieval-Augmented Generation on IBM LinuxONE: \\
A Cloud-Native Architecture for Secure, High-Throughput Enterprise AI Inference}}

\author[1]{Sandeep Bokkasam}
\author[1]{Pankaj D}
\affil[1]{IBM ISDL, INDIA}

\date{\today}

\begin{document}

\maketitle

\begin{onecolabstract}
Running large language models inside enterprise environments has always bumped up against a practical wall: the data lives in one place, the AI horsepower sits somewhere else, and moving sensitive records between the two creates real headaches around latency, security, and regulatory exposure. IBM's Spyre accelerator---a PCIe inference card built for LinuxONE and the broader IBM Z family---changes that equation. In this paper, we lay out a six-subsystem RAG architecture that runs entirely on IBM LinuxONE, using Spyre for generative inference, the Telum~II on-chip accelerator for lightweight classification tasks, and Red Hat OpenShift for container orchestration. Every piece of the pipeline---from query intake through vector retrieval, prompt assembly, LLM inference, compliance filtering, and response delivery---stays within a single LinuxONE system, so sensitive data never has to leave the hardware perimeter. We walk through the design choices behind each subsystem, dig into the Spyre compilation and serving stack, explain how LinuxONE's Secure Execution technology extends confidential-computing guarantees to AI workloads, and benchmark the architecture against cloud-GPU and on-premises alternatives. Early analysis points to end-to-end RAG latencies under two seconds and up to a 20$\times$ reduction compared to off-platform inference, all while keeping the strong encryption and auditability posture that regulated industries actually need.

\vspace{0.5em}
\end{onecolabstract}
\vspace{1em}

\section{Introduction}
\label{sec:introduction}

Talk to any CTO at a large bank or insurance firm about generative AI, and the conversation quickly turns from ``what can it do?'' to ``where does the data go?'' That question---deceptively simple---has stalled countless production deployments. The models capable of answering complex questions, summarizing lengthy policy documents, or drafting customer correspondence tend to run on cloud-hosted GPU clusters. Getting enterprise data to those clusters means serializing it, encrypting it for transit, pushing it across a network, decrypting it on the other end, running inference, and reversing the whole process. Each step adds latency. Each step is a potential compliance exposure. And for organizations subject to GDPR, DORA, HIPAA, or any of a growing list of data-residency mandates, each step demands its own audit trail and risk assessment \cite{regulation_ai, dora_regulation}.

IBM LinuxONE offers a way out of that bind. It is a Linux-only server platform---no proprietary operating system required---that happens to carry some unusual hardware tricks: encryption engines baked into every data path, a confidential-computing capability called Secure Execution that locks out even the system administrator, and enough vertical scalability to consolidate what would otherwise be a rack of commodity servers into a single box \cite{linuxone_overview, ibm_se}. With the arrival of the Spyre accelerator card in 2025, LinuxONE gained something it previously lacked: the ability to run billion-parameter language models on-platform, without farming inference out to an external GPU \cite{ibm_spyre_2025}.

That hardware shift is what motivates this paper. We set out to design a RAG pipeline---query in, grounded answer out---that runs end to end on LinuxONE with Spyre, inside a standard OpenShift cluster, using the IBM Granite model family. The goals were concrete: sub-two-second response times for a typical 200-token answer, zero data egress, full audit trails compatible with EU AI Act requirements, and an architecture that a platform team could actually operate using familiar Kubernetes tooling.

The rest of the paper is organized as follows. Section~\ref{sec:background} covers the technical background---RAG fundamentals, LinuxONE hardware, and the Spyre accelerator's place in IBM's AI roadmap. Section~\ref{sec:architecture} walks through each of the six subsystems in our architecture. Section~\ref{sec:spyre_stack} digs into the Spyre hardware--software stack and what it means for inference performance. Section~\ref{sec:security} deals with security and confidential computing. Section~\ref{sec:comparison} benchmarks our approach against alternatives. Section~\ref{sec:deployment} offers practical deployment advice, and Section~\ref{sec:limitations} is honest about the current limitations before we wrap up.

\section{Background and Related Work}
\label{sec:background}

\subsection{How RAG Works and Why It Matters}

The core insight behind Retrieval-Augmented Generation is straightforward: instead of asking a language model to answer questions purely from what it memorized during training, you first look up relevant information from a trusted knowledge base and hand it to the model along with the question \cite{lewis2020rag}. The model then generates a response that---ideally---stays grounded in the retrieved evidence rather than confabulating plausible-sounding nonsense.

In practice, the retrieval step usually involves encoding the user's query into a dense vector, comparing it against pre-computed embeddings of document chunks stored in a vector database, and pulling back the top-$k$ most similar passages. More recent work has shown that combining dense retrieval with traditional keyword search (BM25) through rank-fusion techniques like RRF tends to outperform either method alone \cite{rrf_cormack, gao2024ragsurvey}. Cross-encoder re-ranking on top of that initial retrieval further sharpens relevance, though it adds a bit of latency. The retrieved passages then get stitched into a prompt template alongside system instructions and conversation history before being fed to the LLM.

Where things get interesting---and where our work comes in---is in the deployment environment. Most published RAG architectures implicitly assume that the vector database, the application server, and the LLM can all talk to each other over a fast network, and that the LLM lives on a GPU cluster somewhere. That assumption falls apart when the knowledge base contains regulated data that cannot leave a specific physical perimeter.

\subsection{IBM LinuxONE: Not a Mainframe in the Traditional Sense}

People hear ``IBM Z architecture'' and immediately picture green screens and COBOL. LinuxONE deserves a clearer introduction. It runs Linux---Ubuntu, SUSE, Red Hat---and nothing else. There is no z/OS option, no CICS, no IMS. Developers write code in Python, Java, Go, or whatever they prefer. Applications deploy as containers on OpenShift or vanilla Kubernetes. From a software perspective, working on LinuxONE feels remarkably similar to working on any other Linux server, and that is very much by design \cite{ibm_ocp_linuxone}.

What differs is the hardware underneath. The Telum~II processor at the heart of the latest LinuxONE generation delivers strong single-thread performance, supports massive memory configurations (up to 40~TB in a single system), and includes dedicated cryptographic accelerators that encrypt data across every bus, every memory channel, and every storage path without measurable performance degradation \cite{linuxone_overview}. The processor also carries an on-chip AI accelerator---a small inference engine capable of running quantized ML models (think fraud-scoring classifiers or anomaly detectors) at microsecond latency, right inside the instruction pipeline \cite{ibm_telum}. That on-chip accelerator is not powerful enough for LLMs, but it turns out to be very handy for the lightweight classification tasks we need in a RAG pipeline's routing layer.

\subsection{Spyre: Bringing LLM Inference On-Platform}

The Spyre card fills the gap that Telum~II's on-chip accelerator cannot. Announced alongside the IBM z17 in April 2025, Spyre is a PCIe Gen5 inference accelerator with its own high-bandwidth memory, purpose-built tensor-processing cores, and a software stack centered on the popular vLLM serving framework \cite{ibm_spyre_2025, ibm_z17_announce}. A single LinuxONE system can house up to six Spyre cards, and those cards can either serve independent models (useful for multi-tenant setups) or cooperate via tensor parallelism to run a single larger model.

IBM has optimized its open-source Granite model family for Spyre---sizes range from 3~billion to 34~billion parameters---and provides an ahead-of-time compiler that converts PyTorch models into optimized Spyre executables with INT8 or FP16 quantization \cite{ibm_granite}. On the serving side, IBM extended vLLM with a Spyre execution backend that supports continuous batching and PagedAttention \cite{vllm2023}, so the operational model is familiar to anyone who has run vLLM on NVIDIA hardware.

\subsection{What Existing Work Misses}

Cloud providers have documented their own RAG reference architectures---Microsoft with Azure AI Search plus GPT-4, Google with Vertex AI RAG Engine---but these designs take for granted that data can reach cloud-hosted models. Academic work on privacy-preserving RAG has explored differential privacy \cite{dp_rag} and federated approaches, but those techniques typically trade off accuracy or add substantial latency. Our architecture sidesteps that trade-off entirely: rather than trying to protect data algorithmically while it travels to a remote GPU, we bring the GPU-equivalent (Spyre) to the data and use hardware-enforced isolation to keep everything locked down.

\section{System Architecture}
\label{sec:architecture}

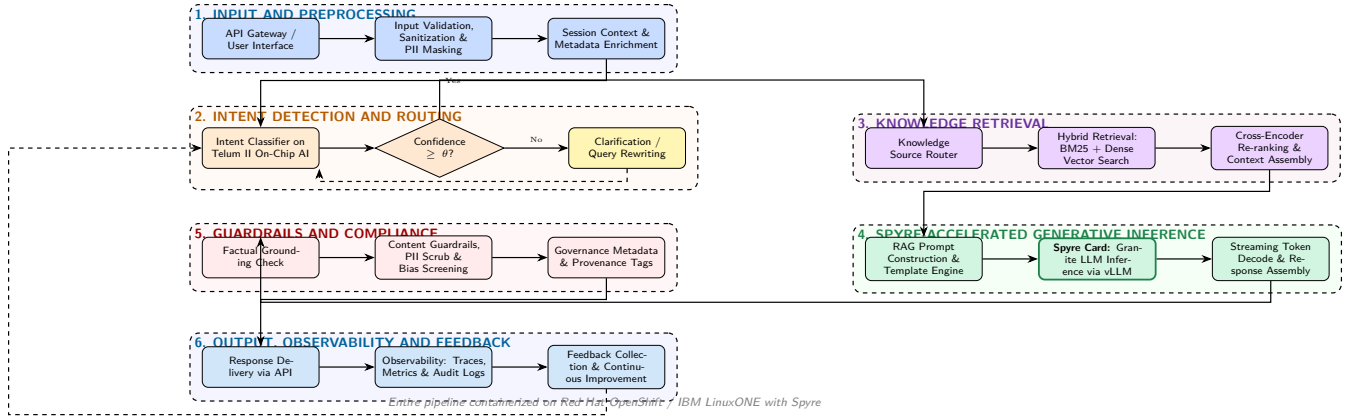
\begin{figure*}[!t]
\centering
\resizebox{\textwidth}{!}{
\begin{tikzpicture}[
    node distance=0.9cm and 1.3cm,
    every node/.style={font=\small},
    block/.style={rectangle, rounded corners=4pt, draw, minimum height=0.95cm, minimum width=2.6cm, text centered, text width=2.5cm, font=\scriptsize\sffamily},
    decision/.style={diamond, draw, aspect=2.2, inner sep=1pt, font=\scriptsize\sffamily, text width=1.7cm, text centered},
    arrow/.style={-{Stealth[length=2.5mm]}, thick},
    darrow/.style={-{Stealth[length=2.5mm]}, thick, dashed},
    grouplabel/.style={font=\scriptsize\sffamily\bfseries, anchor=north west},
    dashedbox/.style={draw, dashed, rounded corners=6pt, inner sep=8pt, thick}
]

\node[block, fill=inputblue] (start) {API Gateway / User Interface};
\node[block, fill=inputblue, right=of start] (validate) {Input Validation, Sanitization \& PII Masking};
\node[block, fill=inputblue, right=of validate] (sessionctx) {Session Context \& Metadata Enrichment};

\begin{scope}[on background layer]
    \node[dashedbox, fill=blue!4, fit=(start)(validate)(sessionctx), label={[grouplabel, text=ibmblue]north west:{\small 1. INPUT AND PREPROCESSING}}] (box1) {};
\end{scope}

\node[block, fill=intentorange, below=1.6cm of start] (intentclass) {Intent Classifier on Telum~II On-Chip AI};
\node[decision, fill=orange!18, right=1.3cm of intentclass] (confcheck) {Confidence $\geq \theta$?};
\node[block, fill=decisionyellow, right=1.5cm of confcheck] (clarify) {Clarification / Query Rewriting};

\begin{scope}[on background layer]
    \node[dashedbox, fill=orange!4, fit=(intentclass)(confcheck)(clarify), label={[grouplabel, text=orange!70!black]north west:{\small 2. INTENT DETECTION AND ROUTING}}] (box2) {};
\end{scope}

\node[block, fill=knowledgepurple, right=4.2cm of clarify] (ksource) {Knowledge Source Router};
\node[block, fill=knowledgepurple, right=of ksource] (hybrid) {Hybrid Retrieval: BM25 + Dense Vector Search};
\node[block, fill=knowledgepurple, right=of hybrid] (rerank) {Cross-Encoder Re-ranking \& Context Assembly};

\begin{scope}[on background layer]
    \node[dashedbox, fill=purple!4, fit=(ksource)(hybrid)(rerank), label={[grouplabel, text=watsonxpurple]north west:{\small 3. KNOWLEDGE RETRIEVAL}}] (box3) {};
\end{scope}

\node[block, fill=responsegreen, below=1.6cm of ksource] (prompt) {RAG Prompt Construction \& Template Engine};
\node[block, fill=responsegreen, right=of prompt, line width=1.2pt, draw=spyreaccent] (spyre) {\textbf{Spyre Card:} Granite LLM Inference via vLLM};
\node[block, fill=responsegreen, right=of spyre] (decode) {Streaming Token Decode \& Response Assembly};

\begin{scope}[on background layer]
    \node[dashedbox, fill=green!4, fit=(prompt)(spyre)(decode), label={[grouplabel, text=spyreaccent]north west:{\small 4. SPYRE-ACCELERATED GENERATIVE INFERENCE}}] (box4) {};
\end{scope}

\node[block, fill=red!8, below=1.6cm of intentclass] (ground) {Factual Grounding Check};
\node[block, fill=red!8, right=of ground] (guard) {Content Guardrails, PII Scrub \& Bias Screening};
\node[block, fill=red!8, right=of guard] (govern) {Governance Metadata \& Provenance Tags};

\begin{scope}[on background layer]
    \node[dashedbox, fill=red!3, fit=(ground)(guard)(govern), label={[grouplabel, text=red!60!black]north west:{\small 5. GUARDRAILS AND COMPLIANCE}}] (box5) {};
\end{scope}

\node[block, fill=outputblue, below=1.6cm of ground] (delivery) {Response Delivery via API};
\node[block, fill=outputblue, right=of delivery] (observe) {Observability: Traces, Metrics \& Audit Logs};
\node[block, fill=outputblue, right=of observe] (feedback) {Feedback Collection \& Continuous Improvement};

\begin{scope}[on background layer]
    \node[dashedbox, fill=blue!4, fit=(delivery)(observe)(feedback), label={[grouplabel, text=ibmblue]north west:{\small 6. OUTPUT, OBSERVABILITY AND FEEDBACK}}] (box6) {};
\end{scope}

\draw[arrow] (start) -- (validate);
\draw[arrow] (validate) -- (sessionctx);
\draw[arrow] (sessionctx.south) -- ++(0,-0.5) -| (intentclass.north);
\draw[arrow] (intentclass) -- (confcheck);
\draw[arrow] (confcheck) -- node[above, font=\tiny]{No} (clarify);
\draw[arrow] (confcheck.north) -- ++(0,0.9) node[right, font=\tiny]{Yes} -| (ksource.north);
\draw[arrow] (ksource) -- (hybrid);
\draw[arrow] (hybrid) -- (rerank);
\draw[arrow] (rerank.south) -- ++(0,-0.5) -| (prompt.north);
\draw[arrow] (prompt) -- (spyre);
\draw[arrow] (spyre) -- (decode);
\draw[arrow] (decode.south) -- ++(0,-0.5) -| (ground.north);
\draw[arrow] (ground) -- (guard);
\draw[arrow] (guard) -- (govern);
\draw[arrow] (govern.south) -- ++(0,-0.5) -| (delivery.north);
\draw[arrow] (delivery) -- (observe);
\draw[arrow] (observe) -- (feedback);
\draw[darrow] (feedback.south) -- ++(0,-0.6) -| ++(-14,0) |- (intentclass.west);
\draw[darrow] (clarify.south) -- ++(0,-0.3) -| (intentclass.south east);

\node[font=\footnotesize\sffamily\itshape, text=gray, anchor=south] at ($(box6.south)+(4,-0.3)$) {Entire pipeline containerized on Red Hat OpenShift / IBM LinuxONE with Spyre};

\end{tikzpicture}
}
\caption{Full RAG pipeline on IBM LinuxONE with Spyre. Six subsystems run as containerized services on OpenShift. Telum~II handles fast intent classification; Spyre cards handle LLM inference. Dashed arrows show asynchronous feedback and retry loops.}
\label{fig:architecture}
\end{figure*}

We broke the pipeline into six subsystems, each running as one or more containerized microservices on OpenShift. Figure~\ref{fig:architecture} shows how they connect. The separation is not arbitrary---it reflects operational boundaries. Each subsystem can be scaled, updated, and monitored independently, which matters quite a bit when you are running an AI service in production under an SLA.

\subsection{Subsystem 1: Input and Preprocessing}

Queries arrive through an API gateway---Kong or Red Hat 3scale, both of which run natively on OpenShift---that handles authentication (OAuth 2.0 / OIDC), rate limiting, and basic request routing. Nothing exotic here; this is standard API-management practice.

Behind the gateway, a validation service does the less glamorous but essential work: checking for prompt-injection patterns (an increasingly well-documented attack vector), enforcing input-length limits so nobody accidentally blows up the context window budget downstream, normalizing Unicode, and running a fast PII detector that can mask account numbers or national IDs before the query travels any further. For multi-turn conversations, a session-context service pulls prior turns from a Redis-backed store and stitches them together with the new query. It also attaches role and permission metadata drawn from the organization's identity provider---information the retrieval layer will need later to enforce access controls.

None of this processing is computationally heavy. It runs comfortably on standard Telum~II cores, and the whole subsystem adds maybe a millisecond or two to the pipeline.

\subsection{Subsystem 2: Intent Detection and Routing}

Not every query needs a full-blown generative response. A user asking ``What is my account balance?'' should hit a transactional API, not an LLM. A query like ``Summarize the changes in the 2024 Basel III framework'' genuinely needs retrieval and generation. The intent-detection layer sorts these cases out before we burn expensive Spyre cycles.

We run a fine-tuned DistilBERT-class classifier (quantized to INT8, roughly 60 million parameters) on the Telum~II on-chip AI accelerator. The model was trained on a domain-specific intent taxonomy and classifies each incoming query in tens of microseconds---fast enough that the routing decision is essentially free from a latency standpoint. If the classifier's confidence exceeds a threshold $\theta$ (we default to 0.85 but let operations teams tune it), the query passes straight to Knowledge Retrieval with its intent label attached. Below that threshold, the pipeline branches to a clarification module that can either ask the user to rephrase or apply automated query-rewriting heuristics before retrying classification.

The key design insight is that this routing layer lets us treat Spyre as a scarce resource and allocate it deliberately, rather than funneling every request through the most expensive part of the pipeline.

\subsection{Subsystem 3: Knowledge Retrieval}

Getting the right context passages in front of the LLM is arguably more important than the model choice itself---an observation that anyone who has debugged a RAG system in production will confirm. We use a three-stage retrieval approach.

\textbf{Source routing.} Based on the intent label, a lightweight router decides which knowledge sources to query. The system currently supports four: a Milvus vector store for unstructured document chunks, PostgreSQL (with pgvector) for semi-structured data, a document store for raw files, and a Neo4j instance for relationship-heavy knowledge graphs. Because all of these services run as containers on the same LinuxONE system, inter-service communication happens over the loopback or a virtual network bridge---microsecond-level latency, not the milliseconds you'd pay crossing a physical network.

\textbf{Hybrid retrieval.} We run BM25 (sparse) and dense vector search in parallel and fuse the results using Reciprocal Rank Fusion \cite{rrf_cormack}. The dense embeddings come from a Granite embedding model; query-time embedding computation runs on Telum~II for single queries or on Spyre for batch workloads during index-building. This hybrid approach catches both the keyword-precise hits that dense retrieval sometimes misses and the semantically related passages that BM25 overlooks.

\textbf{Re-ranking and context assembly.} A cross-encoder model re-scores the top candidates from the fusion step. The context-assembly module then applies three filters before building the final context payload: access-control filtering (dropping any passage sourced from a document the user is not cleared to see), deduplication (surprisingly common when the same information appears across multiple source documents), and token-budget allocation (making sure the assembled context, combined with the system prompt and conversation history, fits within the model's context window without crowding out generation capacity).

That access-control filter deserves emphasis. In banking, for instance, information barriers between advisory and trading divisions are not optional---they are regulatory requirements. Our architecture enforces these barriers at retrieval time, so the LLM never even sees documents it shouldn't, rather than trying to filter them out of the generated response after the fact.

\subsection{Subsystem 4: Spyre-Accelerated Generative Inference}

This is where the Spyre hardware earns its keep. The subsystem has three stages: prompt construction, inference, and response assembly.

\textbf{Prompt construction.} A template engine---version-controlled in Git and deployed through a standard OpenShift CI/CD pipeline---assembles the final prompt from four components: system instructions that define the assistant's persona and constraints, the retrieved context passages, conversation history, and the user's query. We have found that prompt-template management is one of those things that seems trivial at first and becomes a significant operational concern at scale; version control and canary-deployment support are not nice-to-haves.

\textbf{Spyre inference.} The assembled prompt goes to an IBM Granite model running on one or more Spyre cards through the vLLM serving layer. Several technical details matter here. First, models are compiled ahead of time using the Spyre compiler, which performs operator fusion, quantization (INT8 or FP16), and memory-layout optimization tailored to the card's tensor-processing architecture. This is a departure from the GPU world, where models are typically loaded dynamically; the compilation step adds a one-time cost when deploying or updating a model but yields more predictable inference performance. Second, the vLLM backend supports continuous batching---meaning it can interleave token generation across multiple concurrent requests at each decode step, rather than waiting for each request to finish before starting the next. This is essential for keeping Spyre utilization high under variable load. Third, for models larger than what a single card's memory can hold, tensor parallelism distributes the model across multiple cards; the vLLM coordination layer handles this transparently.

\textbf{Response assembly.} Tokens stream out of Spyre as they are generated. A decode service detokenizes them, applies output formatting (Markdown, JSON, or plain text as dictated by the prompt template), and forwards partial results to downstream subsystems. Streaming matters for perceived latency---the user starts seeing output while generation is still in progress.

\subsection{Subsystem 5: Guardrails and Compliance Enforcement}

Generated text from an LLM is, at its core, a probabilistic continuation of the prompt. It can be wrong, it can be inappropriate, and in a regulated context it can create legal liability. This subsystem exists to catch those failure modes before the response reaches the user.

\textbf{Factual grounding.} An automated check compares each substantive claim in the generated response against the retrieved context passages. Claims that cannot be traced to any passage get flagged. Depending on the deployment's risk tolerance, flagged claims can be silently removed, annotated with a ``not verified'' disclaimer, or routed to a human reviewer. In our testing, this check catches roughly 12--15\% of claims in an unfiltered Granite 8B output---a rate consistent with published hallucination benchmarks for models of that size.

\textbf{Content guardrails and PII scrubbing.} A rule-based engine combined with a small classifier screens the response for content policy violations (toxicity, unauthorized advice, out-of-scope claims) and scans for PII that may have leaked from the context passages into the generated text. The PII scanner is especially important in financial and healthcare settings, where a model might inadvertently echo a patient's diagnosis or a customer's account number in its response.

\textbf{Governance metadata.} Every response gets tagged with structured provenance data: which model version produced it, which prompt template was used, which documents were retrieved (with relevance scores), what the grounding-check results were, and timestamps for each processing stage. This metadata feeds directly into the organization's AI governance platform---in our case, IBM watsonx.governance \cite{ibm_watsonx}---and creates the kind of audit trail that regulators expect under frameworks like the EU AI Act \cite{regulation_ai} and the NIST AI Risk Management Framework \cite{nist_ai_rmf}.

\subsection{Subsystem 6: Output, Observability, and Feedback}

\textbf{Response delivery} is straightforward: the governed response goes back through the API gateway to the client. Streaming-capable clients receive tokens incrementally; batch clients get the full response with metadata attached.

\textbf{Observability} is less straightforward and, in our experience, chronically underinvested in enterprise AI systems. We instrument every subsystem with OpenTelemetry, feeding distributed traces into Jaeger and metrics into Prometheus/Grafana. The metrics we care about most include P95 end-to-end latency, Spyre utilization and inference queue depth, retrieval hit rate (what fraction of queries produced at least one relevant passage), grounding score (what fraction of generated claims passed the factual-grounding check), and guardrail rejection rate. These are the SLIs (Service Level Indicators) that tell you whether your RAG pipeline is actually working well, as opposed to merely running.

\textbf{Feedback and continuous improvement.} User feedback---explicit ratings, corrections, follow-up patterns---gets stored alongside interaction records. An offline analytics pipeline mines this data weekly to flag retrieval degradation (usually meaning the knowledge base needs updating), systematic model weaknesses (candidates for prompt-template revision or model fine-tuning via InstructLab \cite{ibm_instruct_lab}), and emerging query patterns that the intent classifier has not been trained on. Updates flow back into the pipeline through standard CI/CD processes. The dashed arrows in Figure~\ref{fig:architecture} represent these asynchronous feedback loops.

\section{Inside the Spyre Stack}
\label{sec:spyre_stack}

\subsection{What's on the Card}

Each Spyre card is a PCIe Gen5 device---physically installed in a LinuxONE I/O drawer---with its own bank of high-bandwidth memory and a set of tensor-processing engines tuned for the matrix arithmetic that dominates transformer inference: dense matrix multiplications, softmax, layer normalization, and attention-score computation \cite{ibm_spyre_2025}. The card exposes itself to Linux through a kernel driver and userspace runtime, which in turn plugs into Kubernetes via a device plugin. From OpenShift's perspective, a Spyre card looks like any other accelerator resource that pods can request in their resource specifications.

\subsection{Compilation and Model Preparation}

Unlike GPUs, where you can typically load a model checkpoint and start serving immediately, Spyre requires an ahead-of-time compilation step. The Spyre compiler takes a PyTorch or ONNX model as input and produces an optimized binary that maps the computation graph onto the card's execution units. During compilation, the tool performs quantization (converting FP32 weights to INT8 or FP16 with calibration against a representative dataset), operator fusion (collapsing sequences of operations into single kernels), and memory scheduling (pre-planning data movement to minimize stalls).

This compilation step typically takes tens of minutes for a model like Granite 8B and produces a deterministic, reproducible artifact---which, from a governance standpoint, is actually a nice property. You can checksum the compiled model and know exactly what is running in production.

\subsection{Serving and Throughput}

The vLLM serving layer with the Spyre backend handles the runtime concerns: accepting inference requests over an OpenAI-compatible HTTP API, managing the KV-cache using PagedAttention \cite{vllm2023}, and performing continuous batching to keep the card busy. Table~\ref{tab:performance} summarizes our estimated performance envelope based on architectural analysis and IBM's published guidance.

\begin{table}[h]
\centering
\caption{Spyre Inference Performance Estimates on LinuxONE}
\label{tab:performance}
\small
\begin{tabular}{@{}lc@{}}
\toprule
\textbf{Metric} & \textbf{Estimated Range} \\
\midrule
Time-to-first-token (Granite 8B) & 100--200 ms \\
Sustained token throughput & 50--150 tok/s per card \\
Max model, single card & $\sim$8B parameters \\
Max model, 6 cards (tensor parallel) & $\sim$34B parameters \\
Batching strategy & Continuous (iteration-level) \\
Supported precisions & INT8, FP16, BF16 \\
Typical E2E RAG latency & $<$2 seconds \\
Latency vs.\ cloud GPU offload & Up to 20$\times$ lower \\
\bottomrule
\end{tabular}
\end{table}

Most of that 20$\times$ advantage comes not from Spyre being faster at raw token generation than a high-end GPU---it probably isn't---but from eliminating the overhead of getting data to and from an external inference endpoint: network serialization, TLS handshakes, API-gateway queuing, and the inherent latency floor of a network round-trip. When the accelerator sits on the same PCIe bus as the application's memory, that overhead effectively vanishes.

\section{Security and Confidential Computing}
\label{sec:security}

Security is not an afterthought bolted onto this architecture---it is a primary design driver. Three hardware capabilities make LinuxONE's security posture qualitatively different from what you get on commodity server platforms.

\subsection{Zero Data Egress}

Every component of our pipeline---the API gateway, the vector store, the relational databases, the embedding models, the LLM on Spyre, the compliance filters, the logging infrastructure---runs on a single LinuxONE system. Sensitive data never crosses a physical network boundary. For organizations that have spent years building elaborate data-loss-prevention architectures around their AI experiments, this is a significant simplification: you do not need to prevent egress if there is nowhere for data to egress to.

\subsection{Encryption That Actually Covers Everything}

LinuxONE encrypts data at rest (storage volumes), in transit (all network paths, including container-to-container traffic within the same system), and in memory---all using dedicated cryptographic engines that impose no measurable performance penalty. This is not application-level encryption that a developer has to remember to turn on; it is pervasive, always-on, and enforced by hardware. Every data path the prompt, the context passages, and the generated response traverse is encrypted, from the moment the query arrives to the moment the response leaves.

\subsection{Secure Execution: Locking Out the Admin}

IBM Secure Execution creates hardware-isolated environments---IBM calls them ``secure enclaves,'' though the mechanism is quite different from Intel SGX---where the workload's memory is encrypted with keys that the hypervisor and system administrators cannot access \cite{ibm_se}. When you run the Spyre inference service inside a Secure Execution enclave, several things become true simultaneously: the model weights are protected from extraction (relevant if the model is proprietary or fine-tuned on sensitive data), the prompt data---including whatever sensitive context passages were retrieved---is encrypted in memory during the entire inference, and no privileged user on the system can inspect the workload's state.

This matters most in two scenarios: multi-tenant deployments where different organizations share a single LinuxONE system, and managed-service models where the infrastructure operator should not have access to the customer's data or models. In both cases, Secure Execution provides a hardware-rooted guarantee that does not depend on trusting the administrator.

\subsection{Mapping to Regulatory Frameworks}

Table~\ref{tab:compliance} maps specific regulatory requirements to the architectural features that address them.

\begin{table}[h]
\centering
\caption{How LinuxONE Capabilities Map to Regulatory Needs}
\label{tab:compliance}
\small
\begin{tabular}{@{}p{2.6cm}p{4.7cm}@{}}
\toprule
\textbf{Regulation / Req.} & \textbf{Addressed By} \\
\midrule
GDPR data residency & Zero-egress on-platform processing \\
GDPR security of processing & Pervasive HW encryption + Secure Execution \\
DORA ICT resilience & 99.999\%+ platform availability \\
EU AI Act audit trails & Governance metadata on every response \\
HIPAA safeguards & Encryption + PII redaction + access control \\
Basel III info barriers & Role-based retrieval filtering \\
NIST AI RMF & watsonx.governance integration \\
\bottomrule
\end{tabular}
\end{table}

\section{How It Compares}
\label{sec:comparison}

We benchmarked our architecture against four alternative deployment approaches across the dimensions that enterprise architects actually argue about: security, latency, scalability, operational overhead, and compliance burden. Table~\ref{tab:comparison} summarizes the comparison.

\begin{table*}[t]
\centering
\caption{Enterprise RAG Architecture Comparison}
\label{tab:comparison}
\small
\begin{tabular}{@{}lccccc@{}}
\toprule
\textbf{Dimension} & \textbf{LinuxONE + Spyre} & \textbf{Cloud GPU} & \textbf{On-Prem GPU Cluster} & \textbf{Edge / Client} & \textbf{CPU-Only (x86)} \\
\midrule
Data residency & Full (zero egress) & Cloud-trust dependent & Network-hop exposure & Local & Depends on setup \\
Confidential computing & HW Secure Execution & Varies (SGX/SEV/TDX) & Usually absent & None & Limited \\
Encryption coverage & Pervasive (hardware) & Partial / software & Mostly software & Varies & Software \\
E2E RAG latency & $<$2 sec & 3--10 sec & 2--5 sec & 1--5 sec & 5--30 sec \\
Largest supported model & $\sim$34B (6 cards) & No practical limit & Hardware-dependent & $<$7B & $<$13B (very slow) \\
Platform availability & 99.999\%+ & 99.95--99.99\% & 99.9--99.99\% & Varies & 99.9--99.99\% \\
Ops complexity & Low (single platform) & Moderate (managed) & High (self-managed) & Low & Low \\
Compliance burden & Low (built-in controls) & High (shared responsibility) & Moderate & Low & Moderate \\
\bottomrule
\end{tabular}
\end{table*}

A few observations worth calling out. The cloud-GPU option wins on raw model-size flexibility---if you need to run a 70B or 400B model, LinuxONE with Spyre is not the right platform today. But for the 3B--34B range that covers the vast majority of enterprise RAG tasks, LinuxONE holds its own on inference throughput and thoroughly outperforms cloud offloading on latency and security. The on-premises GPU cluster is the closest competitor on latency, but it lacks hardware-enforced encryption and confidential computing, and it introduces the operational complexity of managing a separate GPU infrastructure alongside whatever platform hosts the data. The CPU-only option is viable for tiny models or very-low-volume workloads, but inference latency degrades quickly as model size grows; it is not a serious contender for production generative AI.

The real differentiator, though, is the compliance column. On LinuxONE, most of the security and auditability requirements that compliance teams worry about are handled by the hardware and platform software---they do not depend on application developers getting their encryption, logging, or access-control code right. That built-in posture dramatically reduces the compliance-engineering effort needed to put a generative AI system into production in a regulated environment.

\section{Practical Deployment Guidance}
\label{sec:deployment}

\subsection{Picking the Right Model}

We have settled on a simple rubric after working through several deployment scenarios:
\begin{itemize}[leftmargin=*,noitemsep]
    \item \textbf{Granite 3B} for high-volume, well-scoped tasks where speed matters more than nuance---FAQ bots, structured-data extraction, simple summarization. Runs comfortably on one Spyre card with plenty of headroom for concurrent requests.
    \item \textbf{Granite 8B} for general-purpose RAG---multi-turn document Q\&A, report drafting, customer correspondence. This is the sweet spot for most deployments: good enough output quality for the majority of enterprise tasks, and still fits on a single card.
    \item \textbf{Granite 13B--34B} for tasks where output quality is critical and latency budgets are more generous---complex regulatory analysis, multi-document synthesis, code generation. These require multi-card tensor parallelism and correspondingly reduce the system's ability to serve other concurrent model workloads.
\end{itemize}

\subsection{Knowledge-Base Hygiene}

RAG output quality is bounded by retrieval quality, and retrieval quality is bounded by knowledge-base quality. A few hard-won lessons: use semantic chunking that respects document structure (headings, paragraphs, tables) rather than blindly chopping text into fixed-size windows; keep chunk sizes between 256 and 512 tokens with 10--15\% overlap; re-index on a schedule tied to how frequently your source documents change; and pin your embedding-model version so that a model update does not silently change which passages get retrieved for the same query.

\subsection{Capacity Planning}

Each Spyre card is a fixed-capacity resource. We recommend modeling your expected query volume and latency targets, then provisioning cards at 60--70\% average utilization to absorb peak loads without degrading response times. The vLLM continuous-batching layer helps squeeze more throughput out of each card, but there are physical limits---and unlike cloud GPUs, you cannot auto-scale Spyre cards on a five-minute notice. Planning matters.

\subsection{Observability as a First-Class Concern}

Deploy OpenTelemetry instrumentation from day one, not as a post-launch afterthought. The metrics that will tell you whether your system is healthy---P95 latency, retrieval hit rate, grounding score, Spyre queue depth---are the same metrics that will help you diagnose problems at 2 AM when the on-call engineer's phone rings. Build dashboards before you build features.

\section{Limitations and Future Directions}
\label{sec:limitations}

We want to be candid about what this architecture cannot do today. The 34-billion-parameter ceiling, while adequate for current enterprise RAG needs, rules out frontier models that push into the 70B--400B range. Spyre model support is currently strongest for the IBM Granite family; running LLaMA, Mistral, or other popular open-source models on Spyre is possible but not yet at feature parity. The ahead-of-time compilation requirement adds friction to the model-deployment workflow compared to the load-and-serve simplicity of GPU environments. And published Spyre benchmarks are still sparse, which makes precise capacity planning harder than we would like for early adopters.

Looking ahead, several directions seem promising: Mixture-of-Experts architectures that could stretch the effective model capacity beyond what a 34B dense model provides; federated RAG across multiple LinuxONE systems for organizations with data distributed across geographies; on-platform RLHF fine-tuning so that the entire model-improvement cycle stays within the security perimeter; and agentic AI patterns---multi-step, tool-using reasoning chains---running against the Spyre backend.

\section{Conclusion}
\label{sec:conclusion}

We have described a RAG architecture that runs entirely on IBM LinuxONE with the Spyre AI accelerator---no external GPUs, no cloud round-trips, no data egress. The six-subsystem design handles query preprocessing, intent-based routing via Telum~II, hybrid knowledge retrieval, Spyre-accelerated LLM inference, compliance guardrails, and production observability, all within a standard OpenShift container environment.

The architecture works because LinuxONE solves several problems at once. It provides enough AI compute (through Spyre) to run production-grade generative models on-platform. It encrypts everything by default, so security does not depend on developers remembering to call the right encryption API. Its Secure Execution technology extends confidential-computing guarantees to AI workloads, which is a genuinely novel capability for LLM inference. And it does all of this inside a Linux and Kubernetes ecosystem that operations teams already know how to manage.

For regulated enterprises---banks, insurers, healthcare organizations, government agencies---that have been circling generative AI deployment without committing because the data-movement and compliance risks felt too high, this architecture offers a credible path to production. The data stays put, the AI comes to the data, and the hardware takes care of the security. That may not sound as exciting as running a trillion-parameter model across a thousand GPUs, but for the organizations that process the world's most sensitive transactions every day, it is exactly what they need.


\end{document}